\documentclass[runningheads]{llncs}
\usepackage{graphicx}
\begin{document}
\title{Robust Artificial Intelligence and Robust Human Organizations}
\titlerunning{Robust Human Organizations and AI}
% If the paper title is too long for the running head, you can set
% an abbreviated paper title here
%
\author{Thomas G. Dietterich}
\authorrunning{Dietterich, T. G.}
% First names are abbreviated in the running head.
% If there are more than two authors, 'et al.' is used.
%
\institute{School of EECS, Oregon State University\\
\email{tgd@cs.orst.edu}}
\maketitle              % typeset the header of the contribution
\begin{abstract}
Every AI system is deployed by a human organization. In high risk applications, the combined human plus AI system must function as a high-reliability organization in order to avoid catastrophic errors. This short note reviews the properties of high-reliability organizations and draws implications for the development of AI technology and the safe application of that technology.
\end{abstract}

The more powerful technology becomes, the more it magnifies design errors and human failures. An angry man who has only his fists, cannot hurt very many people. But the same man with a machine gun can kill hundreds in just a few minutes. Emerging technologies under the name of “artificial intelligence” are likely to provide many new opportunities to observe this “fault magnification” phenomenon.  As society contemplates deploying AI in self-driving cars, in surgical robots, in police activities, in managing critical infrastructure, and in weapon systems, it is creating situations in which errors committed by human users or errors in the software could have catastrophic consequences.

Are these consequences inevitable? In the wake of the Three Mile Island nuclear power plant failure, Charles Perrow published his book “Normal Accidents” (1984) in which he argued that in any sufficiently complex system, with sufficiently many feedback loops, catastrophic accidents are “normal”—that is, they cannot be avoided. 

Partly in reaction to this, Todd LaPorte, Gene Rochlin, Karlene Roberts and their collaborators and students launched a series of studies of how organizations that operate high-risk technology manage to avoid accidents.  They studied organizations that operate nuclear power plants, aircraft carriers, and the electrical power grid. They summarized their findings in terms of five attributes of what they call High Reliability Organizations (HROs; Weick, Sutcliffe, and Obstfeld, 1999):

\begin{enumerate}
\item Preoccupation with failure. HROs believe that there exist new failure modes that they have not yet observed. These failure modes are rare, so it is impossible to learn from experience how to handle them. Consequently, HROs study all known failures carefully, they study near misses, and they treat the absence of failure as a sign that they are not being sufficiently vigilant in looking for flaws. HROs encourage the reporting of all mistakes and anomalies. 
\item Reluctance to simplify interpretations. HROs cultivate a diverse ensemble of expertise so that multiple interpretations can be generated for any observed event. They adopt many forms of checks and balances and perform frequent adversarial reviews. They hire people with non-traditional training, perform job rotations, and engage in repeated retraining. To deal with the conflicts that arise from multiple interpretations, they hire and value people for their interpersonal skills as much as for their technical knowledge. 
\item Sensitivity to operations. HROs maintain at all times a small group of people who have deep situational awareness. This group constantly checks whether the observed behavior of the system is the result of its known inputs or whether there might be other forces at work.
\item Commitment to resilience. Teams practice managing surprise. They practice recombining existing actions and procedures in novel ways in order to attain high skill at improvisation. They practice the rapid formation of ad hoc teams to improvise solutions to novel problems.
\item Under-specification of structures. HROs empower every team member to make decisions related to his/her expertise. Any person can raise an alarm and halt operations. When anomalies or near misses arise, their descriptions are propagated throughout the organization, rather than following a fixed reporting path, in the hopes that the person with the right expertise will see them. Power is delegated to operation personal, but management is completely available at all times.
\end{enumerate}
Paul Scharre, in his book {\it Army of None} (2018), reports that the US Navy adopted HRO practices on Aegis cruisers after the Vicennes incident in which the Vincennes, an Aegis cruiser, accidentally shot down an Iranian civilian airliner resulting in the death of all passengers and crew. The Aegis system is an autonomous ship defense system. Scharre suggests that the safe deployment of autonomous weapon systems requires that the organization using the systems be a high-reliability organization.
There are at least three lessons for the development and application of AI technology. First, our goal should be to create combined human-machine systems that become high-reliability organizations. We should consider how AI systems can incorporate the five principles listed above. Our AI systems should continuously monitor their own behavior, the behavior of the human team, and the behavior of the environment to check for anomalies, near misses, and unanticipated side effects of actions. Our AI systems should be built of ensembles of diverse models to reduce the risk that any one model contains critical errors. They should incorporate techniques, such as minimizing down-side risk, that confer robustness to model error (Chow, et al., 2015). Our AI systems must support combined human-machine situational awareness, which will require not only excellent user interface design but the creation of AI systems whose structure can be understood and whose behavior can be predicted by the human members of the team. Our AI systems must support combined human-machine improvisational planning. Rather than executing fixed policies, methods that combine real time planning (receding horizon control or model-predictive control) are likely to be better-suited to improvisational planning. Researchers in reinforcement learning should learn from the experience of human-machine interactive planning systems (Bresina and Morris, 2007). Finally, our AI systems should have models of their own expertise and models of the expertise of the human operators so that the systems can route problems to the right humans when needed.

A second lesson from HRO studies is that we should not deploy AI technology in situations where it is impossible for the surrounding human organization to achieve high reliability. Consider, for example, the deployment of face recognition tools by law enforcement. The South Wales police have made public the results of 15 deployments of face recognition technology at public events. Across these 15 deployments, they caught 234 people with outstanding arrest warrants. They also experienced 2,451 false alarms—a false alarm rate of 91.3\% (South Wales Police, 2018). This is typical of many applications of face recognition and fraud detection. To ensure that we achieve 100\% recall of criminals, we must set the detection threshold quite low, which leads to high false alarm rates. While I do not know the details of the South Wales Police procedures, it is easy to imagine that this organization could achieve high reliability through a combination of careful procedures (e.g., human checks of all alarms, looking for patterns and anomalies in the alarms, continuous vetting of the list of outstanding arrest warrants and the provenance of the library face images). But now consider the proposal to incorporate face recognition into the “body cams” worn by police in the US. A single officer engaged in a confrontation with a person believed to be armed would not have the ability to carefully handle false face recognition alarms. It is difficult to imagine an organizational design that would enable an officer engaged in a firefight to properly handle this technology.  

A third lesson is that our AI systems should be continuously monitoring the functioning of the human organization to check for threats to high reliability. As AI technology continues to improve, it should be possible to detect problems such as over-confidence, reduced attention, complacency, inertia, homogeneity, bullheadedness, hubris, headstrong acts, and self-importance in the team. When these are detected, the AI system should be empowered to halt operations.

In summary, as with previous technological advances, AI technology increases the risk that failures in human organizations and actions will be magnified by the technology with devastating consequences. To avoid such catastrophic failures, the combined human and AI organization must achieve high reliability. Work on high-reliability organizations suggests important directions for both technological development and policymaking. It is critical that we fund and pursue these research directions immediately and that we only deploy AI technology in organizations that maintain high reliability.

%
%
%

%
% ---- Bibliography ----
%
% BibTeX users should specify bibliography style 'splncs04'.
% References will then be sorted and formatted in the correct style.
%
% \bibliographystyle{splncs04}
% \bibliography{mybibliography}
%

\end{document}